\documentclass[11pt,twocolumn]{scrartcl}

\usepackage{casa_conf}	
\usepackage{graphicx}	
\usepackage{comment}
\usepackage{multirow}
\usepackage{subfigure}
\usepackage{hyphenat}

\title{Computer Vision for Wildlife Monitoring: Detecting Brown Howler Monkeys using YOLO}

\author{Gabriel Ferri Schneider\\
       PUCRS\\
       gabriel.ferri@edu.pucrs.br
       \and
       Guido Luis Glufke Mainardi\\
       PUCRS\\
       guido.mainardi@edu.pucrs.br
       \and
       Paulo Ricardo Knob\\
       PUCRS\\
       paulo.knob@edu.pucrs.br
       \and
       Patrícia Dias\\
       UERGS\\
       patricia-dias@uergs.edu.br
       \and
       Márcia Jardim\\
       SEMA\\
       marcia-jardim@sema.rs.gov.br
       \and
       Júlio César	Bicca-Marques\\
       PUCRS\\
       jcbicca@pucrs.br
       \and
       Soraia Raupp Musse\\
       PUCRS\\
       soraia.musse@pucrs.br
}

\begin{document}

\maketitle

\begin{abstract}
Urban expansion threatens global biodiversity, especially affecting arboreal species due to the fragmentation of forest habitats. The movement of arboreal species across disjointed forest patches increases mortality risk and, thus, compromises their conservation. In this context, the installation of canopy bridges can be a viable strategy; yet continuous monitoring of their use by arboreal species is essential for ensuring their effectiveness, typically carried out with the aid of camera traps. 
However, this method often produces false-positive images that demand time from conservationists for review. In this context, computer vision algorithms can optimize the task of detecting target species using the canopy bridges. 
In this study, we explored the automatic detection of brown howler monkeys \textit{(Alouatta guariba)} in videos obtained by camera traps. Given the need for a large number of annotated images of the target animals to train the algorithms, we tested the incorporation of auxiliary data to improve detection models,
fine-tuning the YOLOv10 framework using varying proportions of them.
The improvement of these automatic detection techniques contributes to conservation efforts, by providing automatic tools to monitor solutions that minimize the impact of human interference in animals habitats.
\end{abstract}
\linebreak
\linebreak


\section{Introduction}

Habitat fragmentation driven by urban expansion and deforestation poses a critical threat to global biodiversity, isolating wildlife in diminishing forest remnants~\cite{Chaves2022-mj,Sijtsma2020-sf}. For arboreal species, the loss of canopy connectivity increases the risk of mortality from road accidents and electrocution, while simultaneously restricting gene flow and access to resources~\cite{Lala2022-wd,villalobos2022wildlife}. In this context, canopy bridges have emerged as a practical mitigation strategy, providing artificial pathways that facilitate safe movement across fragmented landscapes~\cite{Gregory2013-Carrasco-Rueda,Yap2022-fr,Gregory2022-xi}.

The evaluation of these conservation measures relies heavily on continuous monitoring, typically conducted via indirect observation and camera traps~\cite{Flatt2022,Teixeira2013-ou,Bhardwaj2022-mk}. Although effective, camera traps produce an overwhelming volume of data, much of which consists of false-positive triggers caused by vegetation movement or weather conditions. The manual analysis of thousands of hours of footage is not only time-consuming but also creates a significant delay in assessing conservation outcomes~\cite{Sebastian2022}.
In this context, Convolutional Neural Networks (CNNs) have demonstrated significant potential in automating the detection and classification of primates~\cite{paulet2024deep,shukla2019primate,Zhang2018Towards}. However, the development of high-performance models is often hindered by the "data bottleneck", it means, the requirement for large, manually annotated datasets~\cite{roh2019survey,alzubaidi2023survey}. Synthetic data generation through computer graphics (CG) offers a scalable alternative, allowing for the creation of diverse, pre-labeled training sets, that can supplement limited real-world data.

This study explores a hybrid approach, integrating real-world footage with synthetic data generated in Unity\footnote{https://unity.com/}, to improve the detection of the brown howler monkey (\textit{Alouatta guariba}). We utilize the YOLOv10~\cite{yolov10} architecture to quantify the performance gains from data mixing and propose an automated triage protocol designed to facilitate the analysis of wildlife monitoring videos. By reducing the manual workload, this research contributes to more efficient and scalable conservation strategies for arboreal mammals in urbanized environments.

\section{Related Work}

Vogg et al.~\cite{vogg2025computer} conducted a survey on the use of computer vision algorithms for primate detection, which included the use of synthetic data for augmenting training datasets.
Schieber et al.~\cite{schieber2024indoor} presented a survey on the most used methods for synthetic data generation. In their work, the authors categorized the methods in four groups: Crop-out, Graphics API, 3D game engines, and 3D modeling. 
Fabbri et al.~\cite{fabbri2021motsynth}
used a game engine to generate synthetic data for pedestrian detection and achieved state-of-the-art performance using models trained exclusively on synthetic data. Montanha and Musse~\cite{montanha2025exploring} fine-tuned a CNN for pedestrian detection using a mix of real and synthetic data generated in a game engine, finding that synthetic data can improve model performance and robustness.

Zhang et al.~\cite{Zhang2018Towards} proposed a monkey face detector using a R-CNN and a dataset with over 20,000 monkey faces. Their work had two interesting findings. First, the Viola-Jones face detection algorithm does not work so well for monkey faces, when compared with human faces, especially because of lips and eyebrows. Second, the used R-CNN worked pretty well for monkey face detection, but required
a large number of training samples. Additionally, the authors comment that a pre-training using human
faces was useful in mitigating the problem of shortage of monkey faces for the training stage.
Xu et al.~\cite{xu2018face} investigate the problem of facial detection for golden monkeys from China. The proposed approach consists of three main stages: localization of the monkeys’ bodies, identification of candidate facial skin regions, and detection of actual faces. Experimental results show that the proposed method can reliably localize monkey bodies across images of different sizes and accurately detect their faces. 
Pineda et al.~\cite{pineda2021evaluation} proposed a model to detect Japanese macaque, while also evaluating the effect of transfer learning in the detection accuracy of the models. The authors comment that the best performance was achieved using YOLOv4, while transfer learning increased mean average precision and reduced training convergence time.

Shukla et al.~\cite{shukla2019primate} motivate their work by principles of human perception, introducing a representation learning strategy that is resilient to nuisance variations while preserving identity-related similarities across individual-specific varied traits. Their method is named Primate Face Identification (PFID), and is implemented by training a network to discriminate between matching and non-matching image pairs. Experimental evaluations demonstrate state-of-the-art performance in facial recognition for rhesus macaque (\textit{Macaca mulatta}) and chimpanzees (\textit{Pan troglodytes}) across four evaluation protocols: classification, verification, closed-set identification, and open-set recognition. Paulet et al.~\cite{paulet2024deep} present an initial investigation into the design of a non-invasive deep learning–based framework for facial detection, and individual identification, of Japanese macaques (\textit{Macaca fuscata}). The long-term objective is to leverage identity annotations from the dataset to enable the automatic construction of social network representations for the observed population. The results highlight the promise of the proposed approach as a valuable tool for monitoring individuals and supporting social network analysis in studies of Japanese macaques.

\section{Methodology}
\label{sec:method}

This study combines real-world footage and synthetic data to overcome the scarcity of annotated data in wildlife monitoring, focusing on the automated detection of brown howler monkeys \textit{(Alouatta guariba)}.

\subsection{Primary Dataset}
\label{sec:primarydataset}

\begin{figure}[!htb]
  \centering
  \includegraphics[width=0.45\linewidth]{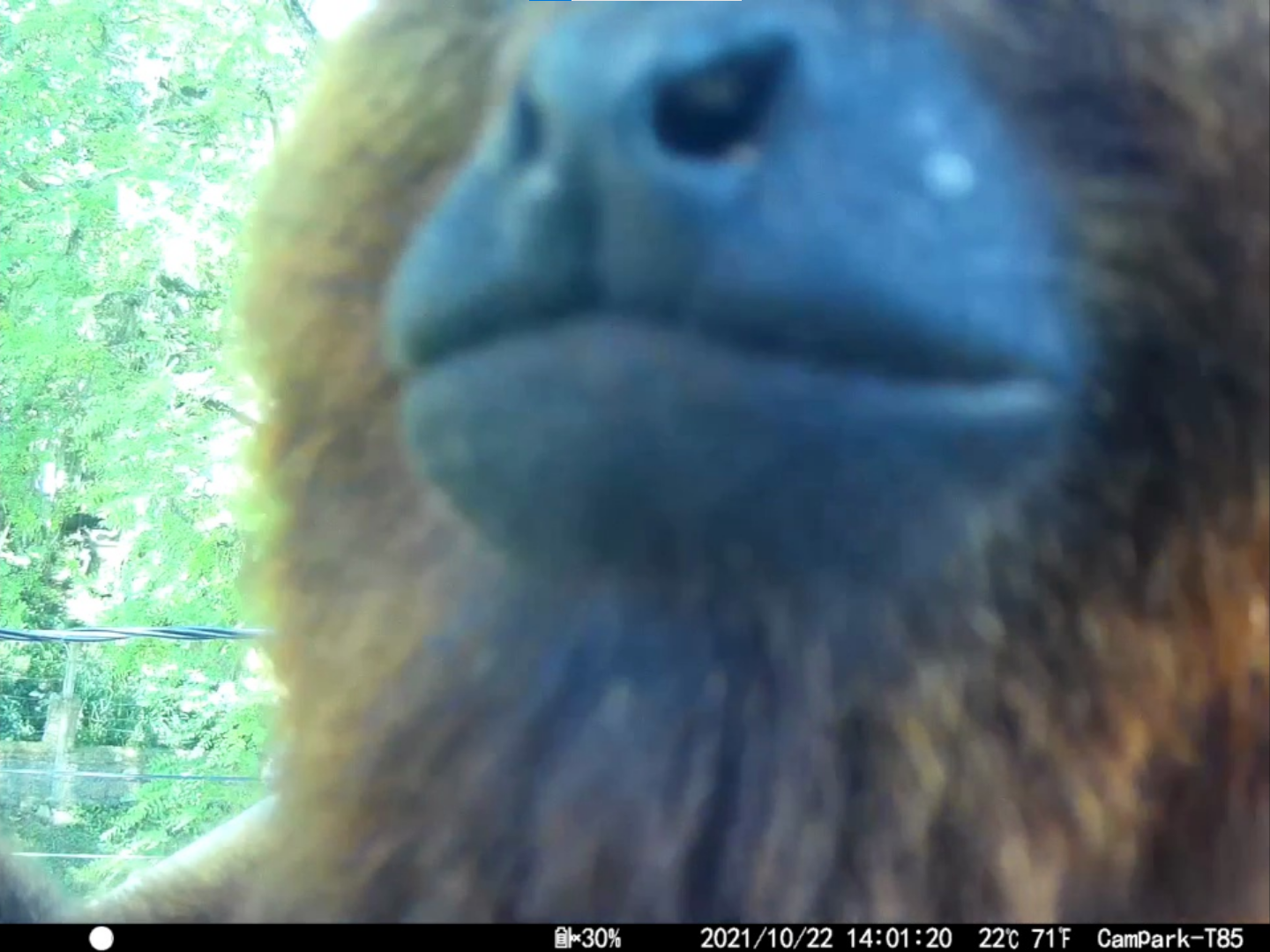}
  \includegraphics[width=0.45\linewidth]{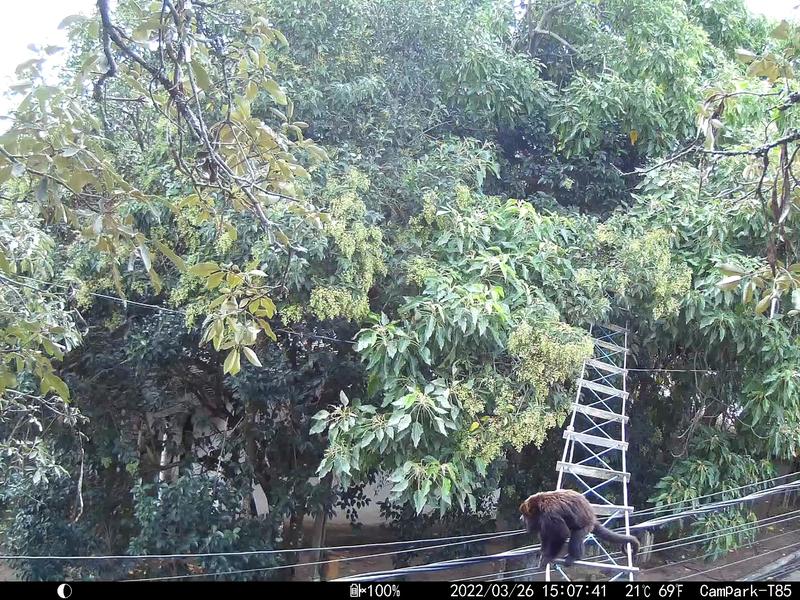}
  \includegraphics[width=0.45\linewidth]{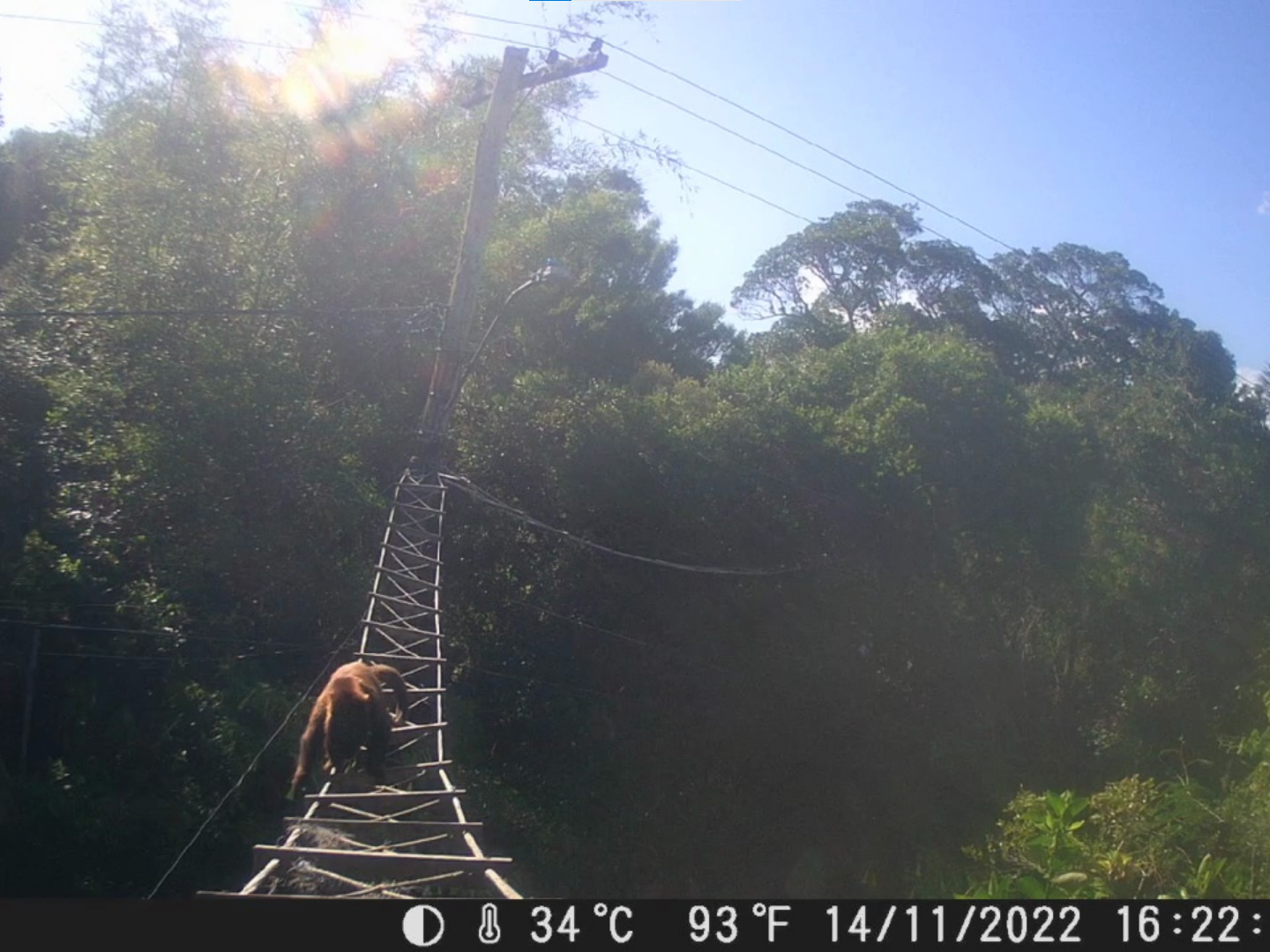}
  \includegraphics[width=0.45\linewidth]{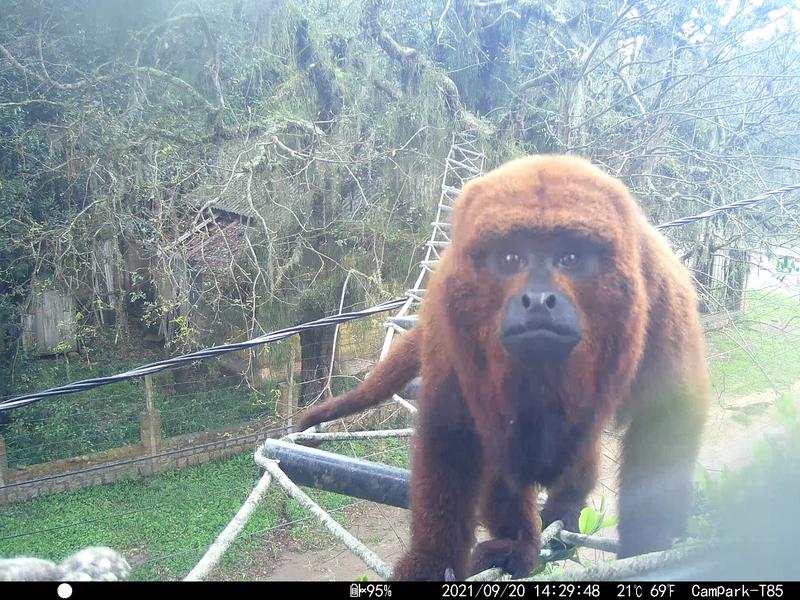}
  \caption{Image examples from the Primary Dataset.}
  \label{fig:bugiosreal}
\end{figure}

The study conducted by Dias~\cite{dias2025behavior} employed systematic field monitoring of arboreal mammals in peri-urban habitats, emphasizing brown howler monkeys (\textit{Alouatta guariba}) using canopy bridges in the Lami neighborhood of Porto Alegre, RS, Brazil. Ten rope-ladder bridges (12-mm ropes; PVC; or treated-wood rungs), installed since 1999, were evaluated. Motion-triggered cameras (Toguard H85, Campark T85; 20 MP, up to 1296P) were positioned to span each bridge and configured for 30s clips with 10s retrigger intervals, yielding 16,179 videos, 3,201 of which contained mammals and 3,026 independent events ($\geq$20-min inter-recording interval). Figure~\ref{fig:bugiosreal} shows example frames for these recordings.

Alongside howlers, recordings included porcupines (\textit{Coendou spinosus}), opossums (\textit{Didelphis albiventris}), and unidentified small rodents; standardized field notes on timing, environmental conditions, and group size complemented the imagery. Thus, a subset comprised by 2,325 videos, in which brown howler monkeys were found, was curated by removing segments where no howlers were present and concatenating the remaining footage per bridge, totalizing about 3 hours. From these compilations, frames were sampled at 1 second intervals to maximize temporal diversity and minimize near-duplicates, producing 10,508 images containing howlers. To prevent temporal leakage, images were partitioned into non-overlapping subsets (90\% training, 10\% testing). All images were manually annotated with the VGG Image Annotator~\cite{Dutta2019VIA} for supervised detection task.

\subsection{Auxiliary Datasets}
\label{sec:auxiliardataset}

\begin{figure}[!htb]
  \centering
  \includegraphics[width=0.45\linewidth]{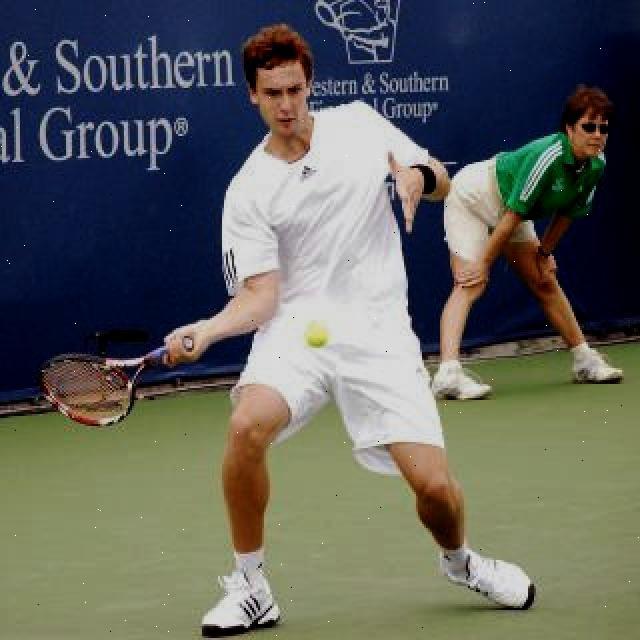}
  \includegraphics[width=0.45\linewidth]{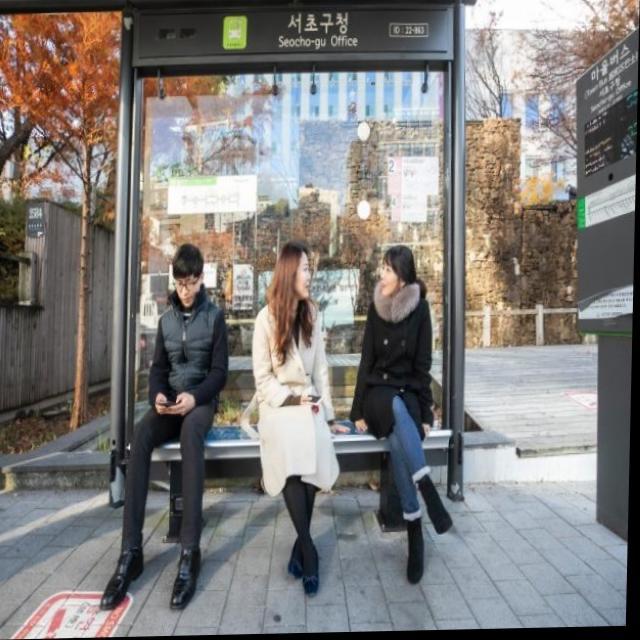}
  \includegraphics[width=0.45\linewidth]{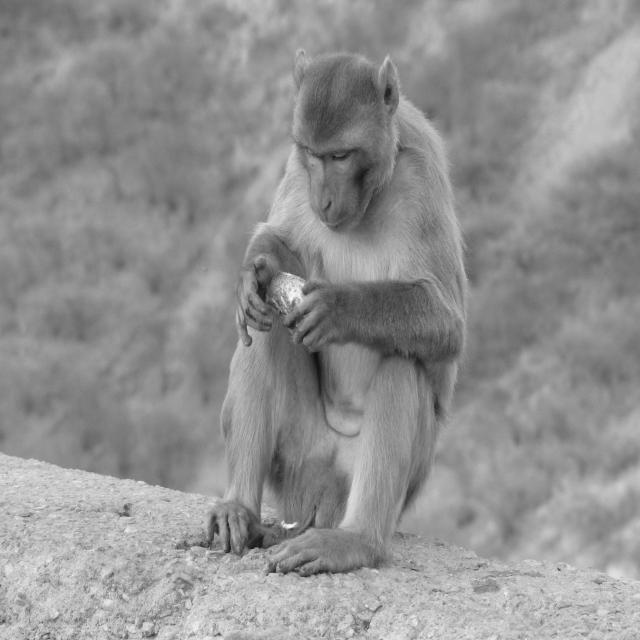}
  \includegraphics[width=0.45\linewidth]{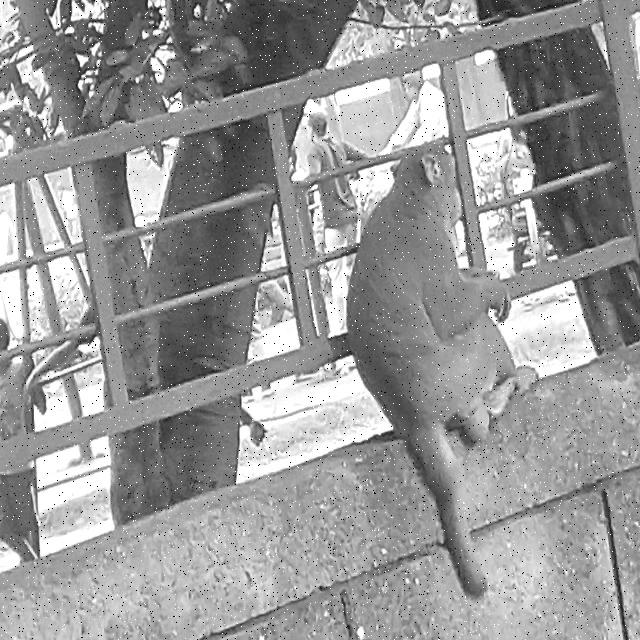}
  \caption{Image examples from the Auxiliary Datasets. On the top row, two examples from “Human Detection Dataset”. On the bottom row, two examples from ``Non-Human Primate Dataset".}
  \label{fig:auxdata}
\end{figure}

In order to explore the use of diversity in the network training,
besides our primary dataset, we obtained two additional public datasets from Roboflow Universe\footnote{https://roboflow.com/universe}.
The first is the “Human Detection Dataset”, which is comprised of 5,000 images containing people, annotated for object detection. The second is the “Non-Human Primate Dataset” and contains 5,000 images of diverse primate species in multiple contexts, also annotated for detection. Figure~\ref{fig:auxdata} shows example images from both datasets.

\begin{figure}[!htb]
  \centering
  \includegraphics[width=0.9\linewidth]{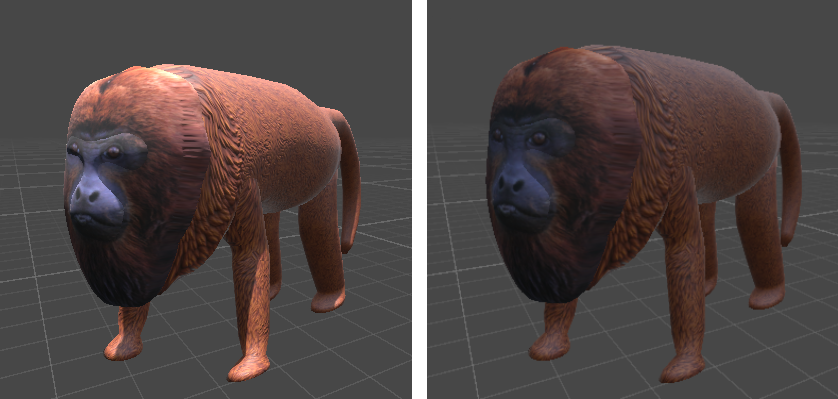}
  \caption{Brown howler monkey model used to generate the synthetic dataset, under two different lighting configurations.}
  \label{fig:bugio}
\end{figure}

\begin{figure}[!htb]
  \centering
  \includegraphics[width=0.45\linewidth]{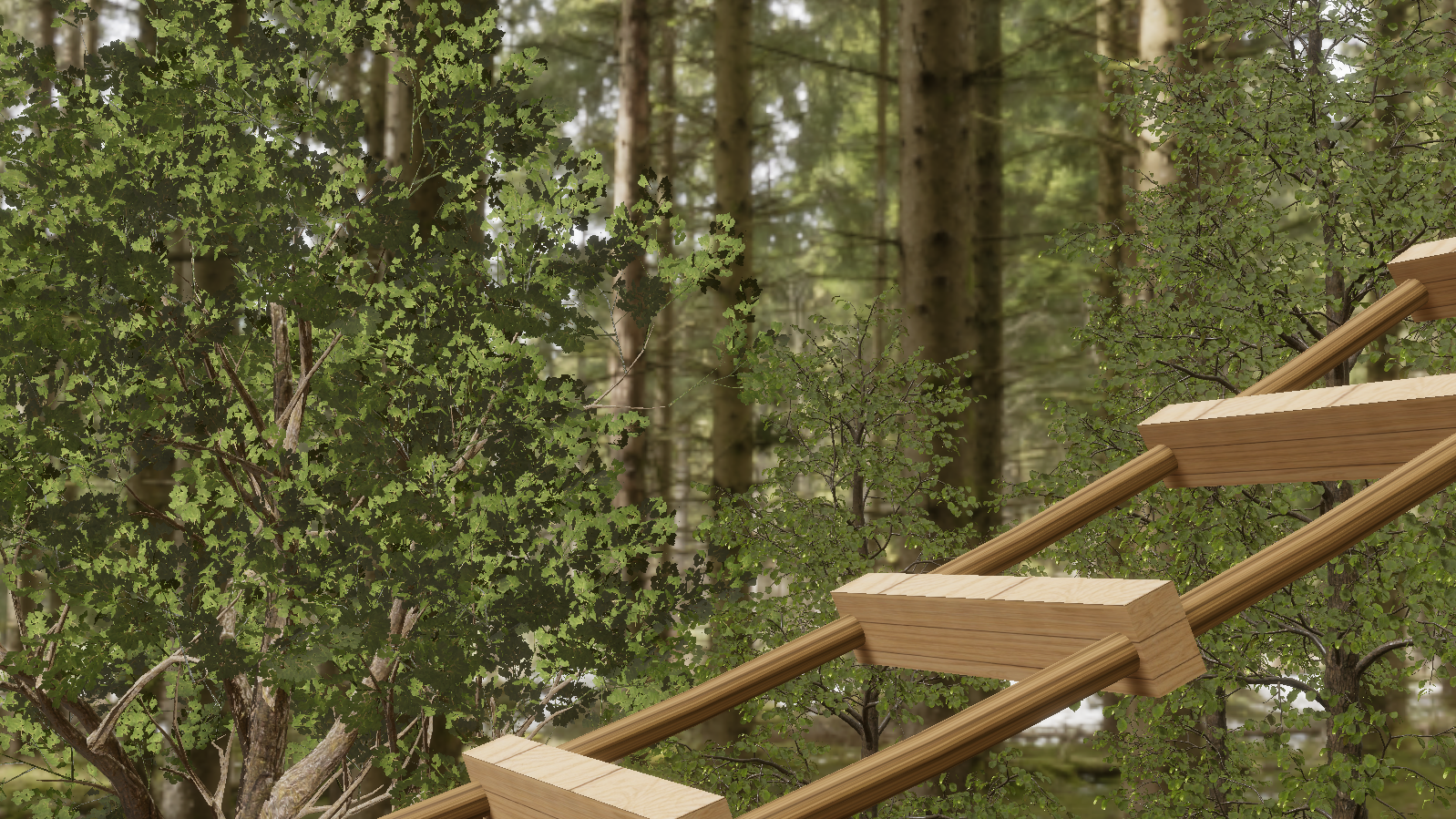}
  \includegraphics[width=0.45\linewidth]{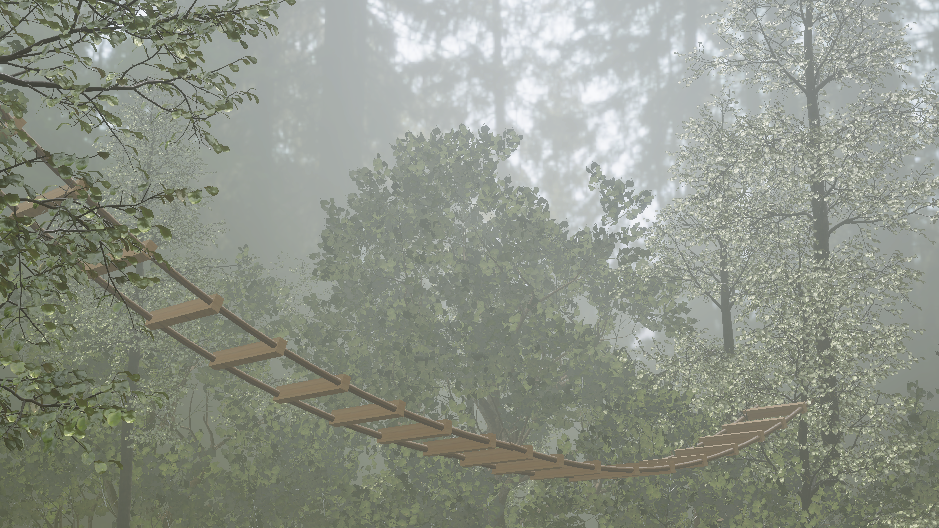}
  \includegraphics[width=0.45\linewidth]{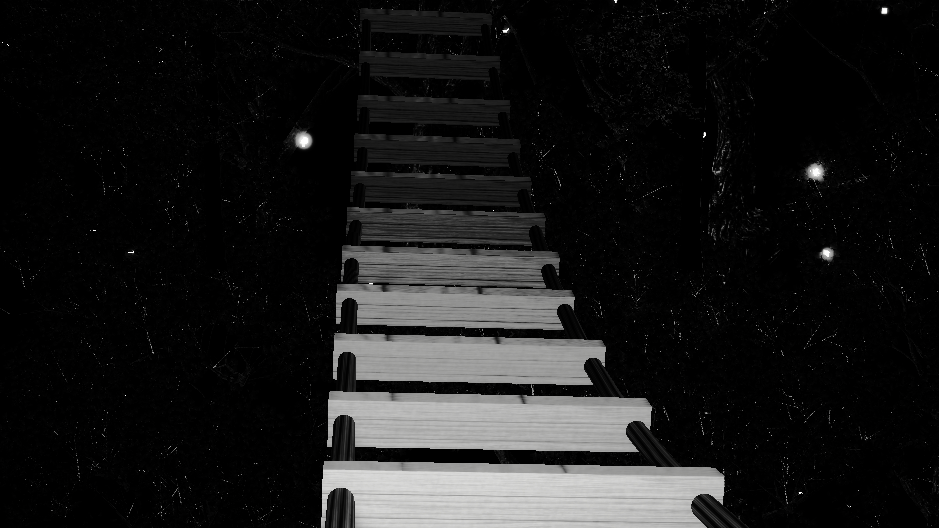}
  \includegraphics[width=0.45\linewidth]{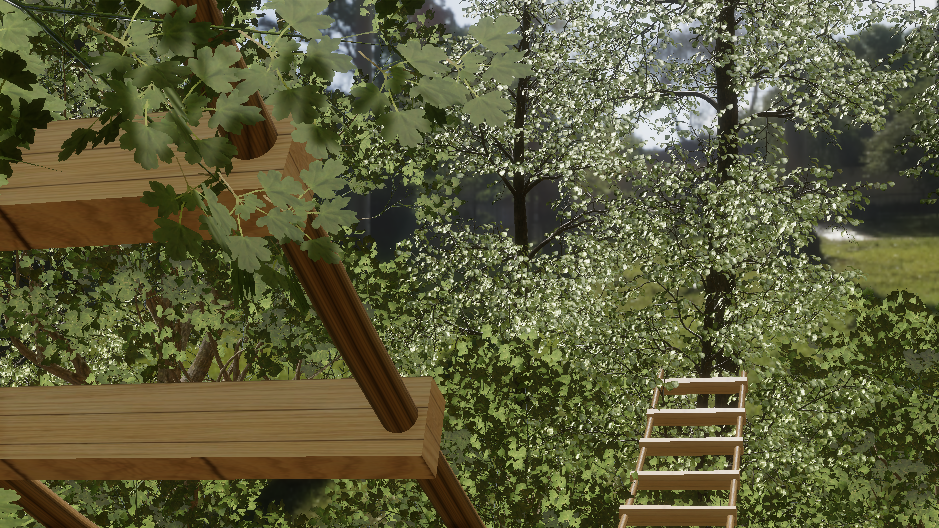}
  \caption{Scenario examples created for the synthetic dataset generation.}
  \label{fig:fundos}
\end{figure}

These auxiliary datasets were used to study whether increased training diversity improves generalization; specifically, whether the model can learn features useful for detecting howlers from related visual contexts. By augmenting the training set with non-target species samples, we seek to test if the model can learn discriminative characteristics relevant to howlers from out-of-domain data. Finally, besides the public datasets, we developed a synthetic dataset using Unity\footnote{https://unity.com} to increase environmental diversity and test the impact of synthetic data on fine-tuning. The development process was as follows:

\begin{itemize}
    \item \textbf{3D model development:} A fully textured 3D brown howler monkey model was created in Blender\footnote{https://www.blender.org}. A real brown howler monkey facial photo was used in Monster Mash~\cite{dvorovzvnak2020monster} to generate the head, while the body was manually modeled and textured. Figure~\ref{fig:bugio} shows the model used with two different lighting configurations.

    \item \textbf{Scene setup and Lighting:} In Unity, we created a scene containing a canopy bridge and some trees, obtained from the internet. For lighting, 9 High Dynamic Range Images (HDRIs) from PolyHaven\footnote{https://polyhaven.com} were used as light maps and background images. In addition, 3 post-processing settings were created to simulate broad daylight, cloudy weather, and nighttime, totaling 27 lighting combinations. Figure~\ref{fig:fundos} presents examples of those background configurations.
    
    \item \textbf{Synthetic dataset construction:} We generated 10,000 synthetic images, each containing one to three howlers placed in one of the 27 total lighting combinations. Nighttime images accounted for 5\% of the dataset, while the remaining 95\% was equally split between clear and cloudy weather, as to match the observed distribution in the primary dataset. The camera position was varied between each shot to obtain different angles and distances.
    
\end{itemize}

\subsection{Video Triage}
\label{sec:videoscreening}

In order to evaluate the model's ability to filter out videos that do not contain brown howler monkeys or other useful information, without discarding those in which the animals appear, we designed a procedure to assess the model’s ability to triage videos.
Starting from the 16,179 camera-trap videos (Section~\ref{sec:primarydataset}), we systematically sampled one out of every three videos, forming a set of approximately 5,000 videos.
Each video is 30s long. To convert the videos to images, we extracted four frames per second, yielding 120 images per video (30s × 4 fps).
As reference for evaluation, all sampled videos were previously annotated by Dias~\cite{dias2025behavior} through manual inspection and categorized as either (i) “contains howler”, if at least one howler appeared at any point, or (ii) “does not contain howler”, if none appeared. These annotations served as ground truth for evaluating the model’s ability to correctly identify and retain videos with howlers, while discarding negatives.

\subsection{Metrics and Evaluation}

We evaluated detection performance using the following standard metrics for classification: Precision, Recall and F1-score. Besides that, we calculate the Intersection over Union (IoU), which quantifies the overlap between the position found by the model and the actual position of the object. IoU is used to calculate Mean Average Precision at IoU 0.5 (mAP@0.5), which is a primary metric in object detection used to evaluate how well a model identifies and locates objects. In our experiments, we prioritize F1-score and mAP@0.5 for evaluating the detection quality.

We adopted temporal cross-validation for the evaluation of our model, because the data was collected over time and, thus, have a strong temporal component associated. This avoids training on frames, or videos, recorded after the validation split, preventing data leakage and yielding more realistic performance estimates. As discussed in the literature~\cite{cerqueira2020evaluating}, conventional cross-validation with random partitioning is inappropriate for temporally structured data, as it can mix future instances into the training set. Thus, we do the following:

\begin{itemize}
    \item \textbf{Temporal partitioning (split)}: Instead of random K-fold, we chronologically ordered the data and partitioned it into successive temporal blocks. Each train/validation iteration respects time order.
    \item \textbf{Cross-validation}: We performed K temporal folds, sliding the validation block forward in time. For each iteration, the training used only data preceding the validation block, which was then evaluated on the subsequent block.
    \item \textbf{Averaging}: Metrics were averaged across folds to obtain robust estimates. This approach is recommended when temporal order must be preserved~\cite{cerqueira2020evaluating}.
\end{itemize}

For the model validation, we also reserved a hold-out set comprising the last 10\% of images per bridge. These images were used exclusively for evaluation and were never input to any model during training or hyperparameter selection. This procedure mitigates metric inflation due to overfitting or context-specific advantages.

\section{Experiments}

To evaluate the impact of auxiliary datasets and synthetic data in fine-tuning YOLOv10 for brown howler monkey detection, we performed experiments with
adding varying proportions of the out-of-domain data.
The focus was on quantifying the impact of the data incorporation, both real and synthetic, for use in the scenarios. 
Following that, we tested the effectiveness of the best performing model as a video triage tool for use by conservationists.

\subsection{Image Experiments}
\label{sec:experiments}

We fine-tuned YOLOv10 for 500 epochs, with early-stop enabled, using different combinations of datasets to assess performance in detecting howlers. 
The goal was to determine whether the incorporation of auxiliary data, both real and synthetic, can improve the detection of brown howler monkeys.
The CG dataset was included as an alternative strategy for data scarcity, given that CG sets can be generated with embedded annotations, eliminating manual labeling costs. 
From the primary dataset, we selected 5,000 images in order to match less numerous datasets (i.e., auxiliary datasets, Section~\ref{sec:auxiliardataset}).

The first experiment focused on training solely with the primary set, testing training sizes from 5,000 down to 1,250 images, to quantify performance degradation with fewer real samples. It is shown as Reference in Table~\ref{tab:resultados_conjuntos}. These results serve as a reference for interpreting the effect of adding auxiliary data. In addition, the baseline YOLOv10 performance is also reported for the purpose of comparison.

After that, we then conducted mixture experiments combining the primary and auxiliary datasets (shown as Non-Human Primates, Human and Synthetic Data in Table~\ref{tab:resultados_conjuntos}), at the following proportions:


\begin{itemize}
    \item 4,500 primary + 500 auxiliary
    \item 3,750 primary + 1,250 auxiliary
    \item 2,500 primary + 2,500 auxiliary
    \item 1,250 primary + 3,750 auxiliary
    \item 0 primary + 5,000 auxiliary
\end{itemize}

It is important to note that the goal of these experiments is not improve the performance of the reference model, but to achieve the most similar performance as possible using less manually annotated images.

Finally, we performed an experiment to analyze how different combinations of synthetic and real images affect YOLOv10 performance. For that, we used all 10,508 manually annotated images (Section~\ref{sec:primarydataset}) and 10,000 synthetic images (Section~\ref{sec:auxiliardataset}).
Therefore, we tested:

\begin{itemize}
    \item 10,508 real images only
    \item 5,000 synthetic + 5,000 real
    \item 10,000 synthetic + 5,000 real
    \item 10,000 synthetic + 10,508 real
\end{itemize}

The results of this experiment are presented in Section~\ref{sec:imageResults}.

\subsection{Video Classification Experiments}
\label{sec:videoclassificationexp}

In addition, we proposed a method to evaluate the best performing model, that being the one trained with 10,000 real and 10,000 synthetic images (Table~\ref{tab:resultados_cg}), as an automatic video triage tool to identify which field recordings contained howlers. We extracted frames from the 5,000 used videos and sampled four frames per second from each 30s video, as commented in Section~\ref{sec:videoscreening}. The model was run on all extracted frames, and per-video decisions about the existence of a brown howler monkey were made using a frame-count threshold, since no frame-level ground truth was available. We counted frames with positive detections and labeled a video as “contains howler” if the count exceeded a threshold. We tested thresholds from 1 to 50 frames and selected the value that maximized F1-score. The results of this experiment are presented in Section~\ref{sec:videoResults}.

\section{Results}
\label{sec:results}

This section presents the quantitative performance of the YOLOv10 model across the various training configurations detailed in Section~\ref{sec:experiments}, as well as its capability as a video triage tool, where it could be of the most use for conservationists (Section~\ref{sec:videoclassificationexp}). The evaluation focuses on the model's ability to generalize detection capabilities when trained with hybrid datasets.

\begin{table*}[ht]
\scriptsize
\centering

\begin{tabular}{lcccccc}
\hline
\textbf{Auxiliary Dataset} & 
\textbf{Auxiliary Images} & 
\textbf{Howler Images} & 
\textbf{Precision} & 
\textbf{Recall} & 
\textbf{F1-Score} & 
\textbf{mAP@0.5} \\
\hline
\textit{Baseline (no fine-tuning)} & 0 & 0 & 0.134 & 0.266 & 0.049 & 0.042 \\
\hline 
\multirow{5}{*}{\centering Reference (no auxiliary data)} 
 & 0     & 5,000 & 0.839 & \textbf{0.781} & \textbf{0.808} & \textbf{0.805} \\
 & 0     & 4,500 & \underline{0.828} & \underline{0.765} & \underline{0.795} & 0.775 \\
 & 0     & 3,750 & 0.820 & 0.756 & 0.786 & 0.770 \\
 & 0     & 2,500 & 0.819 & 0.680 & 0.739 & 0.720 \\
 & 0     & 1,250 & 0.661 & 0.565 & 0.609 & 0.591 \\
\hline
\multirow{5}{*}{\centering Non-Human Primates}
 & 500   & 4,500 & \underline{0.850} & 0.761 & \underline{0.803} & \underline{0.802} \\
 & 1,250  & 3,750 & \underline{0.855} & 0.754 & \underline{0.801} & 0.777 \\
 & 2,500  & 2,500 & \underline{0.873} & 0.689 & 0.767 & 0.758 \\
 & 3,750  & 1,250 & 0.776 & 0.621 & 0.689 & 0.674 \\
 & 5,000  & 0    & 0.315 & 0.181 & 0.229 & 0.208 \\
\hline
\multirow{5}{*}{\centering Human}
 & 500   & 4,500 & \underline{0.829} & 0.759 & \underline{0.792} & \underline{0.794} \\
 & 1,250  & 3,750 & \underline{0.829} & \underline{0.770} & \underline{0.798} & 0.779 \\
 & 2,500  & 2,500 & \underline{0.827} & 0.704 & 0.757 & 0.716 \\
 & 3,750  & 1,250 & 0.694 & 0.625 & 0.654 & 0.614 \\
 & 5,000  & 0    & 0.271 & 0.114 & 0.160 & 0.118 \\
\hline
\multirow{5}{*}{\centering Synthetic Data}
 & 500   & 4,500 & \underline{0.833} & \underline{0.769} & \underline{0.800} & \underline{0.800} \\
 & 1,250  & 3,750 & \underline{0.869} & 0.744 & \underline{0.800} & 0.786 \\
 & 2,500  & 2,500 & \underline{\textbf{0.889}} & 0.713 & 0.787 & 0.779 \\
 & 3,750  & 1,250 & 0.799 & 0.658 & 0.721 & 0.713 \\
 & 5,000  & 0    & 0.383 & 0.190 & 0.253 & 0.199 \\
\hline
\end{tabular}
\caption{Model performance with different training datasets. In bold are presented the highest values per performance metric, while underlined values are the ones that outperform the best reference model or fall within a 2\% margin.}
\label{tab:resultados_conjuntos}
\end{table*}

\begin{table*}[ht]
\scriptsize
\centering

\begin{tabular}{lcccccc}
\hline
\textbf{Auxiliary Dataset} & 
\textbf{Auxiliary Images} & 
\textbf{Howler Images} & 
\textbf{Precision} & 
\textbf{Recall} & 
\textbf{F1-Score} & 
\textbf{mAP@0.5} \\
\hline
\multirow{2}{*}{\centering Reference (no auxiliary data)} 
 & 0     & 5,000 & 0.839 & 0.781 & 0.808 & 0.805 \\
 & 0     & 10,508 & \textbf{0.920} & 0.782 & 0.845 & 0.847 \\
\hline
\multirow{3}{*}{\centering Synthetic Data}
 & 5,000   & 5,000 & 0.876 & 0.789 & 0.830 & 0.831 \\
 & 10,000  & 5,000 & 0.898 & 0.798 & 0.844 & 0.837 \\
 & 10,000  & 10,508 & \textbf{0.920} & \textbf{0.806} & \textbf{0.859} & \textbf{0.873} \\
\hline
\end{tabular}

\caption{Results of the expanded tests using different proportions of real and synthetic data.}
\label{tab:resultados_cg}
\end{table*}

\subsection{Image Results}
\label{sec:imageResults}

Table~\ref{tab:resultados_conjuntos} shows the fine-tuned model performance with the different proportions and types of real, auxiliary, and synthetic data. It is noteworthy that the best recall, F1-Score, and mAP@0.5 scores were achieved by the reference model trained with all 5,000 images selected from the primary dataset (Section~\ref{sec:experiments}),
at 0.781, 0.808 and 0.805, respectively. Meanwhile, the best precision was achieved by the 50/50 split with synthetic data, which increased the precision by 5\% compared to the same reference model, totaling 0.889.

When training without any auxiliary data, it is noteworthy that the performance degradation was not linear, with a much larger drop between 2,500 and 1,250 when compared to 5,000 to 2,500, despite both being a 50\% decrease. Furthermore, when replacing real data with the auxiliary data, there was also a decrease in the performance metrics; however, in many cases, the performance reduction was less than 2\%. Such cases are underlined in Table~\ref{tab:resultados_conjuntos}. The metric with the most underlined instances is precision, followed by F1-score, with recall and mAP@0.5 tied in the last. That indicates that the auxiliary data induced more false negatives than false positives, and that it affected the detection quality (mAP@0.5).

In all but 2 cases, those being when going from 10\% to 25\% with the human and synthetic datasets, the increase in auxiliary data proportion led to a decrease in F1-Score. When training with only auxiliary data, the performance was severely reduced; however, all cases still outperformed the base YOLOv10 without fine-tuning. It can be seen in Table~\ref{tab:resultados_conjuntos} that Non-Human Primates and Synthetic Data were the ones that scored closest values to the reference, in both F1-Score and mAP@0.5. In general, Non-Human Primates had the best scores when less auxiliary images were used. On the other hand, when increasing the amount of auxiliary images, Synthetic Data presented better scores, except for mAP@0.5 when using all 5,000 auxiliary images.

\begin{figure}[!htb]
  \centering
  \includegraphics[width=0.9\linewidth]{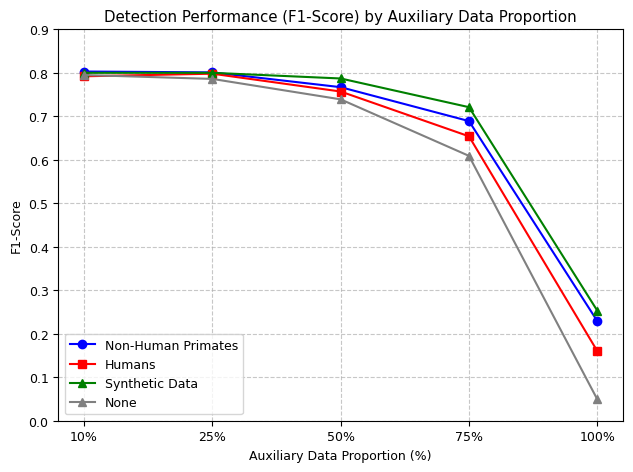}
  \caption{Detection performance of the three variations of auxiliary data, namely Non-Human Primates, Human, and Synthetic Data, for F1-Score.}
  \label{fig:imageResultsF1}
\end{figure}

\begin{figure}[!htb]
  \centering
  \includegraphics[width=0.9\linewidth]{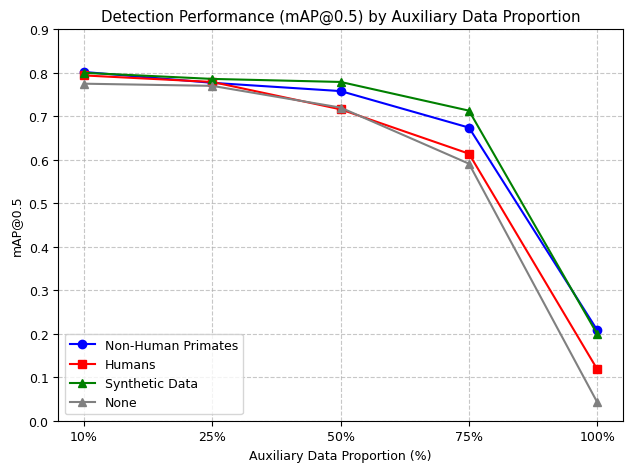}
  \caption{Detection performance of the three variations of auxiliary data, namely Non-Human Primates, Human, and Synthetic Data, for mAP@0.5.}
  \label{fig:imageResultsMap}
\end{figure}

Figures~\ref{fig:imageResultsF1} and~\ref{fig:imageResultsMap} presents the detection performance of the three variations of auxiliary data, namely Non-Human Primates, Human, and Synthetic Data, for both F1-Score and mAP@0.5, respectively.
We calculated the average values of F1-Score and mAP@0.5 for both of them, resulting in: F1-Score - 0.657 (Non-Human Primates) and 0.672 (Synthetic Data); mAP@0.5 - 0.643 (Non-Human Primates) and 0.655 (Synthetic Data).
Overall, considering our prioritized metrics (i.e., F1-score and mAP@0.5), it can be said that the best performing auxiliary dataset was the synthetic dataset, followed by the non-human primate dataset. 

As commented in Section~\ref{sec:experiments}, we also performed an experiment with different combinations of synthetic and real images. The results of this additional fine-tunings, using synthetic data, are presented in Table~\ref{tab:resultados_cg}.
When comparing the 5,000 and 10,000 reference results, it can be seen that precision, F1-score, and mAP@0.5 increased by 9\%, 3.7\% and 4.2\%, respectively, while recall only improved by 0.1\%.
The results training with 5,000 real and 5,000 synthetic images shows a performance that is better in all metrics than training only with 5,000 real images, but worse than training with 10,000 real images only, despite the improved recall. As it can be seen in Table~\ref{tab:resultados_cg}, F1-Score for 5,000 Howler Images (Section~\ref{sec:primarydataset}) plus 5,000 synthetic images (Section~\ref{sec:auxiliardataset}) was 0.830, standing between F1-Scores reached by both trainings without synthetic images. The same pattern occurred with mAP@0.5. This result supports the idea that synthetic data can complement real data, but not fully replace it.

With 5,000 additional (totaling 10,000) synthetic images, making a total of 15,000 images, the model was further improved, leaving it only 0.1\% behind the F1-score from the 10,000 real images, and only 1.0\% behind for mAP@0.5. 
Finally, training with all 20,000 images available, 10,000 real and 10,000 synthetic, resulted in the best model performance, outperforming the 10,000 real images training in both F1-Score and mAP@0.5.

\subsection{Video Classification Results}
\label{sec:videoResults}

For testing the model's potential as a video triage tool, we tested all threshold amounts between 1 and 50 frames, as stated in Section~\ref{sec:videoclassificationexp}. The best performance, as measured by F1-Score, was found using 24 frames as the threshold between positive and negative detection. The performance metrics of the video classification task were the following: 0.762 F1-score, 0.700 precision and 0.838 recall. There is no reported mAP@0.5 score, as the precision of the detection is not of interest in the triage scenario.

We decided to perform a qualitative analysis of the false negative instances, since losing howler detections is more concerning than having a false positive that can easily be removed after manual inspection. When looking at the false negatives, the most common characteristics of these videos were the presence of only one brown howler monkey (89.8\%) and 
the partial appearance of a brown howler monkey (55.5\%), as well as the brown howler monkey only appearing for less than 5 seconds (50.4\%). Environmental factors were also common, with strong sunlight (13.9\%), rain (11.7\%), and nocturnal recordings (9.5\%) being the most present.

These results indicate that the model presents more difficulty when just a few brown howler monkeys appear partially, or for brief periods of time, in the video, especially due to the minimum threshold of positive frames adopted, which tends to discard videos with very short appearances.
Additionally, the results indicate that the task of video classification is more challenging than per-frame classification, with a reduction of 9.7\% in F1-Score compared to the per-frame task. Despite that, the model still showed adequate performance, indicating that it can be reasonably used to aid the work of conservationists. In addition, an app was developed with the intent of easing the use of the trained model and video classification tool by conservationists, which is available in 
Github\footnote{https://github.com/Virtual-Humans-Lab/HowlerApp}.


\section{Final Considerations}

This paper proposed an automated method for detecting brown howler monkeys in video footage acquired from camera traps placed on canopy bridges, using a CNN trained on a combination of real and synthetic data. The proposed solution is designed to optimize the analysis of large-scale video datasets, reducing the reliance on time-consuming manual inspection and enabling more rapid and well-informed decision-making to support the conservation of arboreal primates in fragmented habitats. The integration of synthetic imagery with real images was shown to be an effective mean of addressing the scarcity of annotated samples, a constraint that frequently hinders or limits the practical use of computer vision methods.

Concerning the synthetic images, careful consideration must be given to its realism. While the adopted methodology incorporates variability in illumination, viewpoints, and environmental settings, some aspects of natural behavior, such as social interactions, locomotion across branches, and subtle changes in fur appearance across life stages, remain challenging to accurately reproduce in virtual settings. This constraint does not diminish the usefulness of synthetic data; rather, it underscores the importance of complementing it with real-world recordings to ensure consistency with conditions observed in natural habitats. Future work might include a more thoroughly investigation about the impact of synthetic data characteristics in the training results.

In regards to video classification, while it proved to be a more challenging task for the model, the results are still promising. The achieved recall of 0.838 shows that less than 20\% of videos containing Brown Howlers are lost. However, some technical constraints remain, such as the reliance on fixed detection thresholds, which may lead to the dismissal of videos featuring very brief animal appearances. The current model's performance is still impacted by factors like partial animal visibility and the presence of isolated individuals. Future work can test alternative training schemes and network architectures to further refine the accuracy of automated wildlife triage.

Overall, the proposed approach is promising in the context of wild-life monitoring.
Its capacity to reliably identify the presence of brown howler monkeys while maintaining a low false positive rate highlights its suitability for large-scale monitoring initiatives, supporting both the assessment of different canopy bridge designs and a deeper understanding of the contribution of these structures to the conservation of arboreal mammals. In summary, the evaluated hybrid strategy demonstrated effectiveness in assisting the identification of brown howler monkeys in camera trap footage. This approach offers meaningful support for the management of canopy bridges and, by extension, contributes to mitigating the adverse impacts of habitat fragmentation.




\section*{Acknowledgements}
This study was partly financed by the Conselho Nacional de Desenvolvimento Científico e Tecnológico - Brazil (CNPq) and by Conselho Nacional de Desenvolvimento Científico e Tecnológico - Brazil (INCT SIMAI, CNPq \#408330/2024-4). This paper was supported by the Ministry of Science, Technology, and Innovations, with resources from Law No. 8.248, dated October 23, 1991, within the scope of PPI-SOFTEX, coordinated by Softex, and published in the Residência em TIC 02 - Aditivo, Official Gazette 01245.012095/2020-56.


\bibliographystyle{unsrt}
\bibliography{refs}

\end{document}